\pdfoutput=1
\documentclass[11pt]{article}

\usepackage{EMNLP2022}

\usepackage{times}
\usepackage{latexsym}

\usepackage[T1]{fontenc}
\usepackage[utf8]{inputenc}

\usepackage{microtype}

\usepackage{inconsolata}

\usepackage{enumitem}
\usepackage{amsthm}
\newtheorem{example}{Example}
\usepackage{multirow}
\usepackage{booktabs}
\usepackage{tabu}
\usepackage{amsmath}
\usepackage{lipsum}
\usepackage{xcolor}
\usepackage{hyperref}
\usepackage{tikz}
\usetikzlibrary{arrows,positioning,shapes.multipart} 
\tikzset{
	>=stealth',
	punkt/.style={
		rectangle,
		rounded corners,
		draw=black, thick,
		text centered,
		font=\small,
	},
	rect/.style={
		rectangle,
		draw=black, thick,
		text centered,
		font=\small,
	},
	pil/.style={
		->,
		thick,
		shorten <=2pt,
		shorten >=2pt,
		font=\small,
	},
}
\usepackage{tkz-graph}
\usepackage{tablefootnote}

\newcommand{\LL}[2]{{\upshape[$_\mathsf{#2}$ }#1{\upshape]}}
\newcommand{\WDLINK}[1]{\href{https://www.wikidata.org/wiki/#1}{\url{wd:#1}}}

\title{Semantic Framework based Query Generation for Temporal Question Answering over Knowledge Graphs}

\author{Wentao Ding, Hao Chen, Huayu Li, Yuzhong Qu \\
  State Key Laboratory for Novel Software Technology, Nanjing University, China\\
  \texttt{\{wtding, haochen97, lihuayuu\}@smail.nju.edu.cn, yzqu@nju.edu.cn}}

\begin{document}
\maketitle
\begin{abstract}
Answering factual questions with temporal intent over knowledge graphs (temporal KGQA) attracts rising attention in recent years.
In the generation of temporal queries, existing KGQA methods ignore the fact that some intrinsic connections between events can make them temporally related, which may limit their capability.
We systematically analyze the possible interpretation of temporal constraints and conclude the interpretation structures as the Semantic Framework of Temporal Constraints, SF-TCons.
Based on the semantic framework, we propose a temporal question answering method, SF-TQA, which generates query graphs by exploring the relevant facts of mentioned entities, where the exploring process is restricted by SF-TCons. 
Our evaluations show that SF-TQA significantly outperforms existing methods on two benchmarks over different knowledge graphs.
\end{abstract}

\section{Introduction}
With the rapid growth of knowledge graphs, temporal question answering over knowledge graphs (temporal KGQA) is attracting rising attention in recent years~\citep{JiaARSW18,JiaPRW21}. 
In temporal KGQA, a common phenomenon is that questions express temporal relations between events or time expressions, while knowledge graphs describe the facts resulting from each event. Existing methods handle the heterogeneity between natural language and knowledge graph representation in two ways. Some systems express temporal intents by constructing executable queries, some apply time-sensitive neural models to rank candidate answers. Considering that neural models are difficult to characterize the clear boundaries of concepts (e.g., exactly filter all events that occur ``before 2022''), this paper focuses on generating queries that correspond to the meaning of questions.

From the logic perspective, formulated queries are actually logical restrictions about KG facts. The answers to a question is a set of KG objects, each of which satisfies the corresponding logical restrictions. In previous studies (e.g., \citealp{JiaARSW18}), temporal intents are converted into restrictions over KG facts with quantitative time values. 
Example~\ref{eg:qnt} illustrates a typical conversion from a temporal question to such restriction.

\begin{example}\label{eg:qnt}
``Who was the president of the U.S. when John Lennon was shot?''

The corresponding query on Wikidata can be formulated as the following logical restriction:
\begin{align*}
& T_1 = \mathsf{time}(\mathsf{position\_held}(ANS, \mathsf{U.S.\_president}))\\
& \wedge T_2 = \mathsf{time}(\mathsf{Murder\_of\_John\_Lennon})\\
& \wedge \textsc{Overlaps}(T_1, T_2) .
\end{align*}
\end{example}

However, the idea of constructing queries with quantitative restrictions can not exhaust all possible scenarios. As illustrated in Example~\ref{eg:non-q}, facts with time values are not a necessary premise to introduce a temporal relation.

\begin{example}\label{eg:non-q}
``Where was John Lennon standing when he was shot?''

To construct a comparison restriction, we need to enumerate the ``standing'' of J.L. (i.e. all the experiences of his life). 
The enumeration is hard to implement and might introduce errors.\footnote{For example, Wikidata says that J.L.'s ``residence''(\WDLINK{P551}) includes Liverpool and New York, but does not provide the corresponding time duration.} In fact, the temporal intent does not rely on any time value. The two events occur simultaneously just because they are different aspects of the same entity (\href{https://www.wikidata.org/wiki/Q2341090}{\url{wd:Q2341090}}), the murder of John Lennon.
\end{example}

The above example reveals that intrinsic connections can also make events temporally related. We argue that the neglect of such cases may limit the capability of existing methods. Therefore, the possible temporal constraints, especially those that do not rely on explicit time values, need to be specifically studied. The main challenges in concluding such constraints come from the complexity of natural language and the lack of supervision signals. Practical KGQA tasks often provide only question-answer pairs. i.e., the constraints on the relevant facts are unknown. Manually enumerating all possible constraint structures in a huge search space will be cumbersome or even infeasible. Thus, there is a need for a lightweight method to model the various constraints that correspond to possible temporal intents.

Inspired by the basic idea of frame semantics that ``one cannot understand the meaning of a word without access to all the encyclopedic knowledge that relates to that word.''~\citep{fillmore2006frame},
we assume that temporal intents are expressed as certain constraints about corresponding knowledge and could be interpreted by some structures over KG facts. Specifically, the events involved in a temporal constraint should provide certain KG facts, which support a possible interpretation of it.
We conclude the temporal constraints and their corresponding interpretation structures as the \textit{\underline{S}emantic \underline{F}ramework of \underline{T}emporal \underline{Cons}traints, SF-TCons}. SF-TCons describes what kinds of knowledge are needed and how they are composed in the potential interpretations.
It consists of 6 interpretation structures, which will be presented in Section~\ref{sec:SFoTC}.
To the best of our knowledge, SF-TCons is the first work to systematically summarize the interpretation structures for temporal KGQA tasks.

Based on SF-TCons, we propose a semantic-framework-based question answering method, SF-TQA, to convert SF-TCons into executable queries. SF-TQA generates query graphs by exploring the relevant facts of mentioned entities, where the query graph is a graph representation of executable logical queries that resembles subgraphs of KG~\citep{YihCHG15}.
SF-TQA improves the accuracy of query generation by regarding SF-TCons as restrictions in the exploration. SF-TQA firstly evokes possible interpretations of temporal intents according to TimeML~\citep{PustejovskyLBR10} annotations. It then grounds the temporal elements in corresponding interpretation structures by the relevant KG facts. The grounding phase will generate multiple candidate queries, the best candidate will be distinguished by ranking the pairs of questions and serialized queries with a BERT model.

The rest of this paper is organized as follows: Section~\ref{sec:SFoTC} discusses the SF-TCons in detail. Section~\ref{sec:F2QA}~presents SF-TQA. Section~\ref{sec:eval} evaluates the SF-TQA with two benchmarks over different knowledge graphs. Section~\ref{sec:relwork} summarizes the related work. The last section concludes this paper.

\section{Semantic Framework of Temporal Constraints}\label{sec:SFoTC}
As previously introduced, temporal intents reflect constraints on events and time expressions. We argue that what really supports the constraints is the essential knowledge underlying the involved elements. For example, in a comparison like \textit{``before WWI''}, what is needed is its start time \textit{``1914''} rather than the named entity \WDLINK{Q361} in KG.
Therefore, temporal constraints can be interpreted by describing what kind of knowledge is needed and how they are composed. The interpretation structures of the constraints are presented as SF-TCons, the Semantic Framework of Temporal Constraints.

\subsection{Temporal Constraints in Questions}
Depending on whether the constraints concern quantitative attributes of a single event or the relations between events, we classify the temporal constraints as follows.
\begin{description}[wide]
\item[Value Constraints.] The intentions about quantitative values are often expressed with time values or ordinals (e.g., ``\underline{first} president''). They require certain events to have corresponding temporal or ordinal attributes. Thus, they could be denoted as follow.
\begin{align*}
    &\textsc{HasValue}(E_1, T_1)\tag{VC-1},\\
    &\textsc{HasValue}(E_1, O_1)\tag{VC-2},
\end{align*}
where $E, T, O$ denotes events, time expressions and ordinals respectively.
As an example, the intent ``first president'' could be denoted as $\textsc{HasValue}(\textit{``president''}, \textit{``first''})$.
Specifically, temporal interrogatives (e.g., ``\underline{when} did sth. happen?'') are denoted as $\textsc{HasValue}(E_1, T?)$, which declare the existence of the temporal attributes but has no restrict on the specific value.
    
\item[Relation Constraints.] The possible relations between time and events have been well studied in the AI area. We follow TimeML, the most commonly used annotation specification, to model the relation constraints. 
\begin{example}\label{eg:tml}
TimeML-style annotations for the question in Example~\ref{eg:non-q} :\\
{\upshape Where was John Lennon \LL{standing}{Event_1} \LL{when}{Signal_1} he was \LL{shot}{Event_2}?}\\
{\upshape $\langle$TLINK reltype=\textsc{Simultaneous} target=\textsc{Event$_1$} relatedTo=\textsc{Event$_2$} signal=\textsc{Signal$_1$} /$\rangle$}
\end{example}
As illustrated in Example~\ref{eg:tml}, temporal relations are triggered by certain signals (e.g., ``when'') and classified into pre-defined \textsf{reltype}s. For the practical demand of QA tasks, we formalized the relation constraints as
\begin{align*}
    &\textsc{Relation}(\mathcal{T}_R, E_1, T_1)\tag{RC-1},\\
    &\textsc{Relation}(\mathcal{T}_R, E_1, E_2)\tag{RC-2},
\end{align*}
where $\mathcal{T}_R$ denotes the 13 temporal \textit{reltype}s in TimeML~\citep{PustejovskyCISGSKR03}, $E$ and $T$ denotes events and time expressions respectively. The TimeML-style annotation in the example question corresponds the following 
RC-2 constraint: 
\begin{align*}
\textsc{Relation}(\textsc{Simultaneous}, \textit{``standing''}, \textit{``shot''})    
\end{align*}
\end{description}

\subsection{Interpretation Structure for Temporal Constraints}\label{sec:tc}
As previously mentioned, one temporal constraint could be supported by various interpretations. We summarize 6 interpretation structures (IS) according to whether the involved event expressions are intrinsically connected and what connector between them can correspond to the expected meanings.
In order to enhance the generality of the IS as much as possible, we do not restrict the specific semantic representations of involved events, but only focus on the key knowledge that they can provide. The 6 IS are presented as follows.

\begin{description}[wide]
\item[IS-1 Comparison structure]
    \begin{flalign*}
        &\textsc{HasValue}(E_1, T_1)\,|\,\textsc{Relation}(\mathcal{T}_r, E_1, T_1|E_2)\\
        &\Rightarrow \textsc{Compare$\langle \circ, \mathrm{time}(E_1), \mathrm{time}(T_1|E_2)\rangle$}
    \end{flalign*}
    This structure interprets \textbf{VC-1} and \textbf{RC}, where $\circ$ denotes algebraic predicate for time values~\citep{Allen83,JiaARSW18}. 
    Specifically, the predicate $\circ$ is required to be \textsc{Equal} in \textbf{VC-1} and is determined according to the identified type $\mathcal{T}_r$ in \textbf{RC}. This structure supposes that the involved events provides certain time values.
    
    For example, the question: ``\textit{Which movie did Alfred Hitchcock \LL{direct}{Event_1} \LL{in}{Signal_1} \LL{1960}{Time_1}?}'' corresponds to the following constraint and KG facts, where the ``direct'' event provides the value ``1960-10-7''.
    \begin{align*}
        \textsc{Compare}&\langle \textsc{Includes}, \mathrm{time}(\textit{``direct''}), \textit{``1960''}\rangle
    \end{align*}
    \begin{tikzpicture}[auto,]
	\node[punkt] (E1) {\textsf{Alfred\_Hitchcock}};
	\node[punkt, right=2 of E1] (E2) {\textsf{Psycho} \textit{(ANS)}}
	edge[pil, <-] node[above] {\textsf{director} ($E_1$)} (E1.east);
	\node[punkt, below=0.6 of E2] (T1) {\textsf{``1960-10-7''}}
	edge[pil, <-] node[left] {\textsf{in\_time}} (E2.south);
	\node[punkt, left=2.5 of T1] (T2) {\textsf{``1960''} ($T_1$)};
	\path (T2.east)+(1.2,0) node (A1l) {\small $-$\textsc{Includes}$\rightarrow$};
    \end{tikzpicture}

\item[IS-2 Ordering Structure]
    \[
    \textsc{HasValue}(E_1, O_1) \Rightarrow \textsc{Order$\langle \mathrm{attr}(E_1), O_1 \rangle$}
    \]
    This structure interprets \textbf{VC-2} by ordering entities (or facts) that are described by $E_1$. It supposes that $E_1$ describes a common attribute of certain objects to be ordered.
    For example, the question: \textit{``When did Henry the VIII \LL{marry}{Event_1} his \LL{first}{Ordinal_1} wife?''} corresponds to
    \[
        \textsc{Order}\langle \mathrm{attr}(\textit{``marry''}), \textit{``first''} \rangle
    \]
    \begin{tikzpicture}[auto]
	\node[punkt] (E1) {\textsf{Henry\_VIII\_of\_England}};
	\node[below=0.7 of E1] (DOT) {$\dots$};
	\node[punkt,left=0 of DOT] (A1) {\begin{tabular}{c}
			\textsf{Cathe\dots Aragon} \\
			\textit{(T=1506, 1st (\upshape $O_1$))} 
		\end{tabular}}
	edge[pil, <-, bend left=10] node[left] {\textsf{spouse} ($E_1$)} (E1.west);
	\node[punkt,right=0 of DOT] (A2) {\begin{tabular}{c}
			\textsf{Cathe\dots Parr} \\
			\textit{(T=1543, 6th)}
	\end{tabular}}
	edge[pil, <-, bend right=10] node[right] {\textsf{spouse}($E_1$)} (E1.east);
    \path (A1.west |- A1.north)+(-0.1,0.1) node (A1nw) {};
	\path (A2.south -| A2.east)+(+0.1,-0.1) node (A1se) {};
	\path[rounded corners, draw=black, dashed] (A1nw) rectangle (A1se);
    \end{tikzpicture}

\item[IS-3 Direct Query Structure]
    \begin{flalign*}
    &\textsc{HasValue}(E_1, X) \Rightarrow \textsc{Find$\langle \mathrm{ent}(E_1), \mathrm{attr}(X) \rangle$}&
    \end{flalign*}
    In some cases, the expected values are directly represented in KG facts. 
    This structure interprets \textbf{VC} by directly finding the expected value $X$ in certain attributes of some related entity. It supposes that the entity is related to the mentioned event $E_1$. 
    
    For example, the description: \textit{``\dots did the \LL{7th}{Ordinal_1} \LL{Harry Potter book}{Event_1} \dots''} corresponds to the following representation and KG facts, where the entity ``Harry Potter and the Deathly Hallows'' has some attribute with the value ``7''.
    \[
        \textsc{Find}\langle \mathrm{ent}(\textit{``\dots book''}), \mathrm{attr}(\textit{``7th''}) \rangle
    \]
    \begin{tikzpicture}[auto,]
	\node[punkt] (E1) {\textsf{Harry\dots Hallows}};
	\node[right=3 of E1] (b1) {};
	\node[punkt, above=0.1 of b1] (A1) {\textsf{``7''} ($O_1$)}
		edge[pil, <-, bend right=5] node[above] {\textsf{\ldots.series\_ordinal}} (E1.north east);
	\path (E1.east |- A1.north)+(-0.1,0.2) node (A1nw) {};
	\path (A1.south -| A1.east)+(+0.1,-0.1) node (A1se) {};
	\path[rounded corners, draw=black, dashed] (A1nw) rectangle (A1se);	
	\node[punkt, below=0.1 of b1] (A2) {\textsf{H.P.} (\textrm{$E_1$})}
		edge[pil, <-, bend left=5] node[below] {\textsf{part\_of\dots}} (E1.south east);
    \end{tikzpicture}
\end{description}

\begin{description}[wide]
\item[IS-4 Same Entity Structure]
    \begin{flalign*}
        & \textsc{Relation}(\mathcal{T}_r, E_1, E_2)\\
        & \Rightarrow \textsc{SameEntity$\langle e, \mathrm{attr}(E_1), \mathrm{attr}(E_2) \rangle$}
    \end{flalign*}
    This structure interprets \textit{simultaneous} cases of \textbf{RC-2}. It supposes that the events should be attributes of a certain entity $e$.
    
    For example, the previously introduced question \textit{``Where was John Lennon \LL{standing}{Event_1} \LL{when}{Signal_1} he was \LL{shot}{Event_2}?''} corresponds to
    \begin{align*}
    \textsc{SameEntity}\langle  e, &\mathrm{attr}(\textit{``standing''}),\\
    &\mathrm{attr}(\textit{``shot''}) \rangle
    \end{align*}
    \begin{tikzpicture}[auto,]
	\node[punkt] (E1) {\textsf{Murder\_of\_John\_Lennon} ($e$)};
	\node[right=1 of E1] (b1) {};
	\node[punkt, above=0.3 of b1] (A1) {\textsf{John\_Lennon}}
		edge[pil, <-, shorten >=10pt, bend right=10] node[above] {\textsf{murder.of} ($E_2$)} (E1.north);
	\path (E1.south |- A1.north)+(-0.1,0.2) node (A1nw) {};
	\path (A1.south -| A1.east)+(+0.1,-0.1) node (A1se) {};
	\path[rounded corners, draw=black, dashed] (A1nw) rectangle (A1se);	
	\node[punkt, below=0.3 of b1] (A2) {\textsf{The\_Dakota}(\textit{ANS})}
		edge[pil, <-, shorten >=10pt, bend left=10] node[below] {\textsf{location} ($E_1$)} (E1.south);
	\path (E1.south |- A2.north)+(-0.1,0.1) node (A1nw) {};
	\path (A2.south -| A2.east)+(0.1,-0.2) node (A1se) {};
	\path[rounded corners, draw=black, dashed] (A1nw) rectangle (A1se);	
\end{tikzpicture}

\item[IS-5 Part-of Structure]
    \begin{flalign*}
        & \textsc{Relation}(\mathcal{T}_r, E_1, E_2) \Rightarrow \textsc{PartOf$\langle r_\mathrm{p}, E_1, E_2 \rangle$}
    \end{flalign*}
    This structure interprets \textit{including} cases of \textbf{RC-2}. It does not restrict the representation of events $E_1$ and $E_2$ in KG, but requires that their representation must be connected by a relation $r_\mathrm{p}$ which implies ``part-of''.
    
    For example, the question \textit{``What award did Laurence Fishburne \LL{received}{Event_1} \LL{at}{Signal_1} \LL{the 46th Tony Awards}{Event_2}?''} corresponds to the following representation and KG facts, where $E_1$ corresponds to a \textit{statement}\footnote{\href{https://www.wikidata.org/wiki/Help:Statements}{Statement} is a Wikidata format for representing complex items. It can be roughly considered as RDF blank node.} and $E_2$ corresponds to a named entity.
    \[
    \textsc{PartOf}\langle r_\mathrm{p}, \textit{``received''}, \textit{``\dots Awards''} \rangle
    \]
\begin{tikzpicture}[auto,]
	\node[punkt] (E1) {\textsf{Laurence\_Fishburne}};
	\node[punkt, below=1.5 of E1] (A1) {\textsf{Best\_Featured\_Actor}(\textit{ANS})}
		edge[pil, <-] node[auto] {\textsf{award\_received}} (E1.south);
	\path (A1.north)+(0.8,0.3) node (E1l) {$E_1$};
	\node[punkt, above right=0.2 of E1l] {\textsf{46th\_Tony\_Awards} ($E_2$)}
		edge[pil, <-, bend left=15, shorten >=5pt] node[right] {\textsf{\dots subject\_of} ($r_\mathrm{p}$)} (E1l.east);
	\path (A1.west |- E1.north)+(-0.3,0.1) node (E1nw) {};
	\path (A1.south -| A1.east)+(+0.1,-0.1) node (E1se) {};
	\path[rounded corners, draw=black, dashed] (E1nw) rectangle (E1se);
\end{tikzpicture}

\item[IS-6 Sequent Structure]
    \begin{flalign*}
        & \textsc{Relation}(\mathcal{T}_r, E_1, E_2)\\
        & \Rightarrow \textsc{Sequent$\langle r_\mathrm{\prec}, \mathrm{ent}(E_1), \mathrm{ent}(E_2) \rangle$}
    \end{flalign*}
    This structure interprets \textit{before/after} cases of \textbf{RC-2}. It supposes the events make a pair of related entities to be sequential in KG, where the entities are involved in $E_1$ and $E_2$ respectively and they must be connected by a relation $r_\mathrm{\prec}$ which indicates a preceding (or succeeding) relation.
    
    For example, the question \textit{``Who \LL{became}{Event_1} the president \LL{after}{Signal_1} J.F. Kennedy was \LL{shot}{Event_2}?''} corresponds to
    \[
    \textsc{Sequent}\langle r_\mathrm{\prec}, \mathrm{ent}(\textit{``became''}), \mathrm{ent}(\textit{``shot''}) \rangle
    \]

\begin{tikzpicture}[auto,]
	\node[punkt] (E1) {\textsf{John\_F.\_ Kennedy}};
	\node[left=1.5 of E1] (b1) {};
	\node[punkt, above=0.3 of b1] (A1) {\textsf{Assassination\dots Kennedy}}
		edge[pil, ->, shorten >=10pt, bend left=10] node[above right] {\textsf{murder.of} ($E_2$)} (E1.north);
	\node[punkt, below=0.3 of b1] (A2) {\textsf{Lyndon\_B.\_Johnson}(\textit{ANS})}
		edge[pil, ->, shorten >=10pt, bend right=10] node[below right] {\textsf{\dots replaces} ($r_\prec$)} (E1.south);
\end{tikzpicture}
\end{description}

In summary, IS 1 to 3 interpret the temporal constraints via temporal facts with explicit quantitative values. IS 4 to 6 model the intrinsic connections that can make events temporally related.
It is worth noting that SF-TCons only expresses the expected form of corresponding knowledge, how to obtain the specific knowledge is left to the implementation of question-answering systems. 

\section{Semantic-Framework-Based Temporal Question Answering}\label{sec:F2QA}
\begin{figure}[h]
    \centering
    \includegraphics[width=0.7\columnwidth]{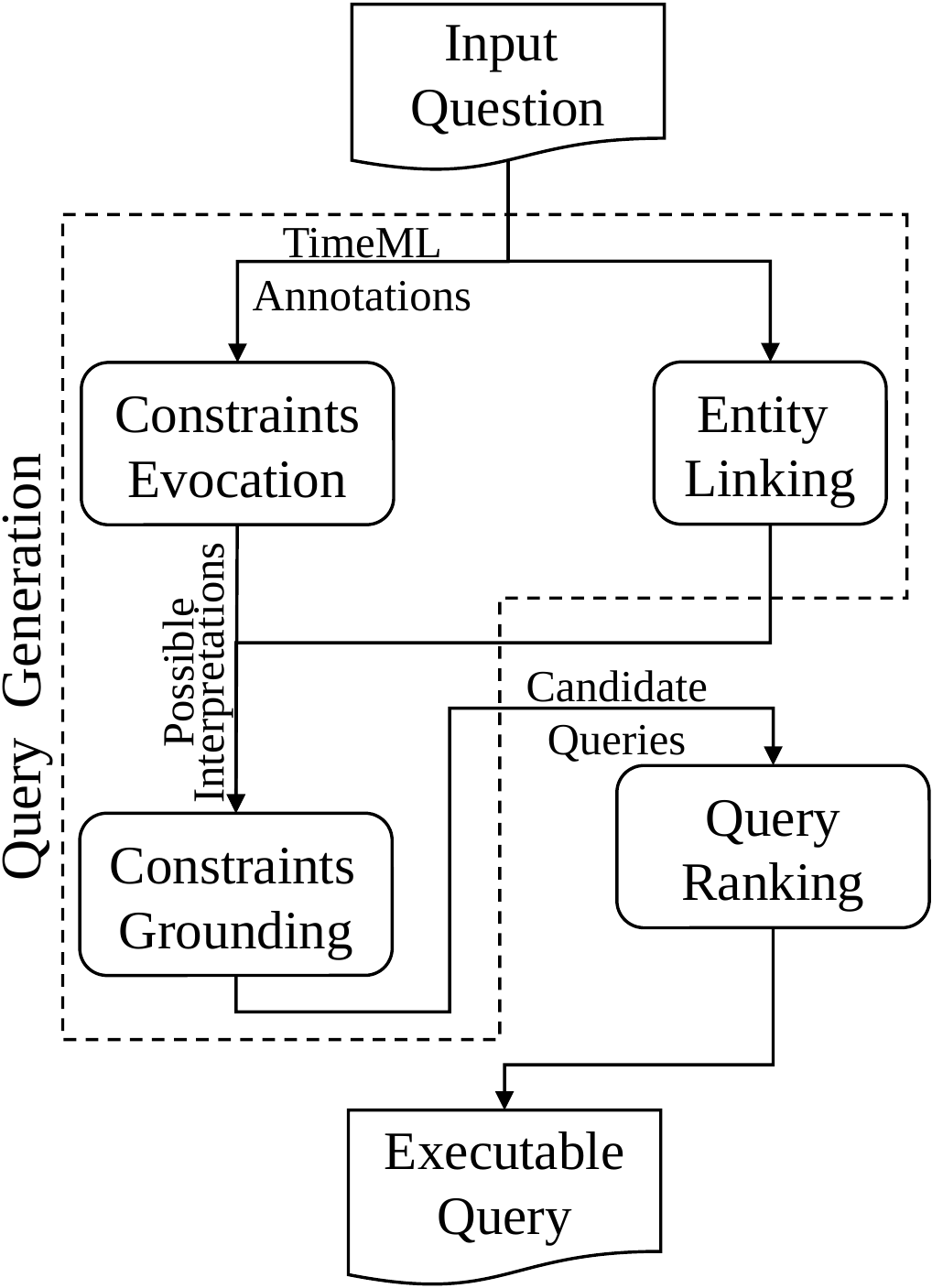}
    \caption{The question-answering process of SF-TQA.}
    \label{fig:frame}
\end{figure}

Figure~\ref{fig:frame} illustrates the question-answering process of the semantic-framework-Based temporal question answering method, SF-TQA. The query generation consists of two steps, 1) evoking the constraints and their possible interpretations (i.e., \textit{constraint evocation}) and 2) grounding the constraints by exploring the relevant KG facts (i.e., \textit{constraint grounding}). The generated candidate queries will be ranked by a BERT model, and the execution results of the highest-scored query will be considered as answers.

\begin{figure*}[ht]
	\centering
	\begin{tikzpicture}[auto]
		\node[punkt, fill=gray, fill opacity=0.3, text opacity=1] (E1) {\textcolor{blue}{\textsf{John\_Lennon}}};
		\node[punkt, below=2 of E1, draw=blue, fill=blue, fill opacity=0.1, text opacity=1] (E2) {\textcolor{blue}{\textsf{Murder\_of\_John\_Lennon}\,($E$)}}
% 		edge[pil, ->, bend right=30] node[below] {\textsf{certain \_aspects\_of}$\dots$} (E1.south)
		edge[pil, ->] node[above] {\textcolor{blue}{\textsf{murder.of}\, $E_2$}} (E1.south);
		\node[punkt, below left=0.5 and 1.5 of E1] (E3) {\textsf{homicide}}
		edge[pil, <-, bend left=5] node[below=-0.1] {\textsf{manner\_of\_death}} (E1.west);
		\node[punkt, below right=0.5 and 1.5 of E1] (E4) {\textsf{ballistic\_trauma}}
		edge[pil, <-, bend right=10] node[below] {\textsf{cause\_of\_death}} (E1.east)
		edge[pil, <-, bend left=10] node[above=-0.2] {\textsf{victim.cause\_of\_death}} (E2.east);
		\node[punkt, bend right=10, above right=0.3 and 2.5 of E1] (T1) {\textsf{``1980-12-08''}}
		edge[pil, <-] node[above] {\textsf{date\_of\_death}} (E1.east);
		\node[punkt, above left=0.3 of E2] (T2) {\textsf{``1980-12-08''}}
		edge[pil, <-] node[auto] {\textsf{time}} (E2.west);
		\node[punkt, above left=0.3 and 1.5 of E1] (T3) {\textsf{``1980''}}
		edge[pil, <-] node[auto] {\textsf{work\_period(end)}} (E1.west);
		\node[punkt, above right=0.2 and 1.5 of E4] (A1) {\textsf{New\_York\_City}}
		edge[pil, <-, bend right=5] node[below=-0.5] {\textsf{place\_of\_death}} (E1.east)
		edge[pil, <-, bend left=15] node[above right] {\textsf{\dots the administrative}} (E2.east);
		\node[punkt, right=4 of E2, fill=gray, fill opacity=0.3, text opacity=1] (A2) {\textcolor{blue}{$ANS$}}
		edge[pil, <-] node[auto] {\textcolor{blue}{\textsf{location}\,$E_1$}} (E2.east);
		\path (T2.west |- T3.north)+(-0.2,0.2) node (B1nw) {};
		\path (E2.south -| A1.east)+(+0.2,-0.4) node (B1se) {};
		\path[draw=black, dashed] (B1nw) rectangle (B1se);
		\node[rect, above right=0.2 and -0.2 of B1nw] (F1) {\textit{``Where was John Lennon when he was shot?''}};
		\node[rect, right=1 of F1] (F2) {$\textsc{SameEntity}\langle  e,\mathrm{attr}(\textit{``standing''}),\mathrm{attr}(\textit{``shot''}) \rangle$};
		\path (F1.east)+(0.4,0) node[] (Dir) {$\Longrightarrow$};
		\path (F2.south)+(0.3,-0.3) node[] (Dir) {$\Downarrow$};
		\path (E1.south west)+(1.1,-1) node[] (A1l) {};
		\path[draw=blue, dashed, fill=blue, fill opacity=0.1] (A1l) ellipse (0.5 and 1);
		\path (E2.east)+(2,0) node[] (A2l) {};
		\path[draw=blue, dashed, fill=blue, fill opacity=0.1] (A2l) ellipse (2 and 0.5);
	\end{tikzpicture}
	\caption{The possible generation process for question in Example~\ref{eg:non-q}. The linked entity and expected answer are colored in \textcolor{gray}{gray} and the facts that can satisfies the evoked interpretation are highlighted in \textcolor{blue}{blue}.}
	\label{fig:eg}
\end{figure*}
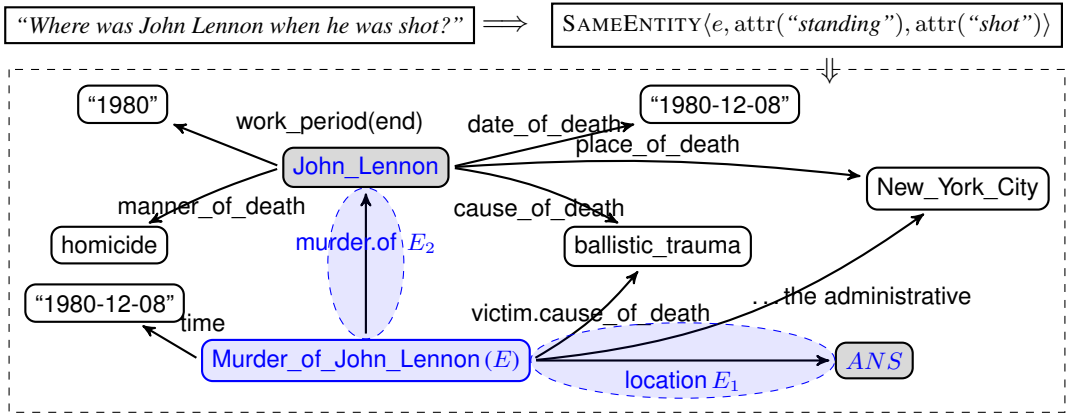

\subsection{Constraint Evocation}
The first step of SF-TQA is to determine the possible constraints. We fine-tune a BERT model to annotate the temporal elements. The corresponding constraints and interpretation structures are evoked according to recognized signals. The elements that involve certain constraints are determined by TimeML relations or by simply taking the temporal elements that are directly described by the signals (i.e., the nearest neighbor of the corresponding signals). The algebraic predicates in the comparison structure are determined by normalizing the TimeML relation types, while other implicit elements are left to the grounding phase. 

\subsection{Constraint Grounding}
In general query graph generation, basic query graphs are constructed as 1 or 2 hop paths from mentioned entities to answers, and they are extended by pre-defined expanding action~\citep{YihCHG15} or fixed interpretation structures of constraints~\citep{BaoDYZZ16}. In temporal KGQA, the main issue is that events could have various representations in KG. As illustrated by the examples in Section~\ref{sec:tc}, they could be represented by named entities, triplet facts, or attributes of their participants. Therefore, we treat the generation of query graphs as grounding the temporal elements in the interpretations of SF-TCons. We divide the descriptions of events into \textit{nominal} and \textit{predicative}. We suppose that nominal descriptions could be the event themselves, and predicative descriptions reflect certain aspects of the events, such as their participants or their post-effect. Therefore, nominal events could be linked entities and others correspond to the neighboring nodes or facts of the explored subgraph(s) or linked entities. The corresponding nodes or facts must provide the knowledge required by corresponding interpolation structures.

We illustrate the above process by the example in Figure~\ref{fig:eg}. In this example, the entity linking module will provide \textsf{John\_Lennon} as a linked entity, and the grounding start with the ``shot'' event which contains the only linked entity. 
We will explore \textit{all} the neighboring facts of \textsf{John\_Lennon} (as illustrated in Figure 2) as candidates for the event. Since ``shot'' is a predicative event and the \textsc{SAME\_ENTITY} constraint requires it to provide an attribute, we will find a triplet that contains \textsf{John\_Lennon} and take the other entity in it (i.e., \textsf{Murder\_of\_John\_Lennon}) as the expected $e$. Similarly, we explore the neighboring facts and select one relation that matches with the question meaning (i.e., \textsf{location} for ``standing''). When there are multiple candidate relations, we will rank the candidates by scoring their serializations with a BERT model. The highest-scored one will be filled in the corresponding slot.

In the specific implementation, which candidates satisfy the question meanings best are determined by neural models. In the training process, we take relations that appear on shortest paths between mentioned entities and answers as positive samples. In particular, the relation that entails \textit{part of} or \textit{precedes} are filtered according to the KG schema in the training process and are predicted by neural models during the test process.
Queries for the questions of multiple constraints are the conjunction of the grounding result of each constraint and queries for the questions with no temporal constraints are unrestricted basic query graphs.

\subsection{Query Ranking}
SF-TQA usually generates multiple candidate queries for one question. We select one of the candidates via neural ranking models. Specifically, we express the generated queries via SPARQL~\footnote{\url{https://www.w3.org/TR/sparql11-query/}} and serialize the queries by dropping auxiliary symbols (e.g., ``\{''). We use the BERT model with cross-entropy loss to score the pair of the input question and serialized queries. 
For each question, we use the candidate queries with the highest $F_1$ score as the positive samples and select $k$ others as negative samples. In order to make our model more robust, we classify the negatives samples as \textit{confusing queries} and \textit{irrelevant queries}. Confusing queries are those that can find partial answers but of lower $F_1$ scores than the positive samples. Irrelevant queries are those whose outputs have no intersection with the correct answers. The ratio of confusing queries to irrelevant queries is $1:1$. The necessity of classifying the negative sample is presented in Appendix~\ref{sec:app}.

\section{Evaluation}\label{sec:eval}
\subsection{Datasets}
We evaluate our method on \textbf{TempQuestions}~\citep{JiaARSW18-2} and \textbf{TimeQuestions}~\citep{JiaPRW21} with the 2015-08-09 dumps of Freebase\footnote{\url{https://developers.google.com/freebase}} and the 2019-01-28 dumps of Wikidata\footnote{\url{https://archive.org/download/wikibase-wikidatawiki-20190128}}, respectively. The statistics of the benchmarks are described in Table~\ref{tab:datasets}. Since TempQuestions does not give the partition between training data and test data, we randomly divided it with the same ratio as TimeQuestions.

\begin{table}[ht]
    \centering
    \begin{tabu} to \columnwidth {X[2, c] X[1, r] X[1, r] X[1, r]}
        \toprule
        \rowfont[c]\bfseries Dataset & Train & Dev. & Test\\
        \midrule
        TempQuestions & 762 & 254 & 255\\
        TimeQuestions & 9,708 & 3,236 & 3,237\\
        \bottomrule
    \end{tabu}
    \caption{The statistics of questions in benchmarks.}
    \label{tab:datasets}
\end{table}

\subsection{Evaluation Metrics}
We report the Hit@1 (denoted as $H@1$), Precision (denoted as $Pr$) , Recall (denoted as $Re$) and $F_1$ score of the evaluation results. Our computation follows~\citeposs{JiaPRW21}\footnote{Their script could be downloaded from \href{https://github.com/zhenjia2017/EXAQT/blob/main/answer_predict/script_listscore.py}{here}.}, where the precision is considered $1$ if the output of a question is empty and the $F_1$ score on a dataset is computed as the average of the scores of each question.

\subsection{Compared Methods}
On TempQuestions, we compare our results with general KGQA methods AQQU~\citep{BastH15}, QUINT~\citep{AbujabalRYW17} and their improved version\citep{JiaARSW18} by incorporating the temporal question decomposition method TEQUILA, QUINT+TEQUILA and AQQU+TEQUILA. On TimeQuestions, we compare our results with general KGQA methods PullNet~\citep{SunBC19}, GraftNet~\citep{SunDZMSC18}, UNIQORN~\citep{Pramanik21} and the temporal KGQA method EXAQT~\citep{JiaPRW21}.

\subsection{Implementation Details}
Our results are obtained on a workstation with an Intel Xeon Gold 5222 CPU, 32 GB of RAM, and NVIDIA RTX3090 GPUs. The hyper-parameters of the ranking models are listed in Table~\ref{tab:hyperp}. They are determined according to the F$_1$ scores on the development sets. We use ELQ~\citep{LiMIMY20} for entity linking. We randomly sample 5,000 questions from the training set of TimeQuestions to fine-tune a BERT model for TimeML annotations. The questions are firstly automatically annotated by simple regex according to their POS-tags and surface forms (e.g., verb tokens may indicate event) then corrected by human annotators. The types of TimeML relations are determined by normalizing the signals via manual rules (e.g., ``during'' corresponds to \textsc{Includes}). 

\begin{table}[hbt]
    \centering
    \begin{tabu} to 0.85\columnwidth {X[2,c] X[1,r] X[1,r]}
        \toprule
        \rowfont[c]\bfseries & TempQ. & TimeQ. \\
        \midrule
        Learning Rate & $5e-5$ & $5e-5$\\
        Batch size & $8$ & $8$\\
        Epochs & $10$ & $15$\\
        Pos./Neg. Ratio & $1:20$ & $1:25$\\
        \bottomrule
    \end{tabu}
    \caption{Hyper parameters for the ranking model.}
    \label{tab:hyperp}
\end{table}

\begin{table}[!h]
    \centering
    \begin{tabu} to \columnwidth {X[2,l] X[1,r] X[1,r] X[1,r] X[1,r]}
        \toprule
        \rowfont[c]\bfseries Method & H@1 & Pr & Re & F$_1$ \\
        \midrule 
        QUINT & 27.0 & 30.3 & \textbf{51.2} & 28.8 \\
        \,+TEQUILA & 31.7 & 40.7 & 42.2 & 32.0 \\
        AQQU & 24.9 & 26.6 & \underline{48.8} & 27.2 \\
        \,+TEQUILA & \underline{36.2} & \underline{40.3} & 43.4 & \underline{37.5} \\
        SF-TQA & \textbf{41.2} & \textbf{42.2} & \underline{48.8} & \textbf{41.1}\\
        \bottomrule
    \end{tabu}
    \caption{Results (\%) on \textbf{TempQuestions}. The best results are in \textbf{bold} and the second bests are \underline{underlined}.}
    \label{tab:TempQ}
\end{table}
\begin{table}[!h]
    \centering
    \begin{tabu} to \columnwidth {X[2,l] X[1,r] X[1,r] X[1,r] X[1,r]}
        \toprule
        \rowfont[c]\bfseries Method & H@1 & Pr & Re & F$_1$ \\
        \midrule 
        PullNet & 10.5 & 5.4 & 19.2 & 7.6\\
        GraftNet & 45.2 & 52.7 & 45.2 & 37.8 \\
        UNIQORN & 33.1 & 14.8 & 45.4 & 19.5\\
        EXAQT & \textbf{56.5} & \textbf{59.3} & \underline{56.9} & \underline{45.6} \\
        SF-TQA & \underline{53.9} & \underline{55.1} & \textbf{62.1} & \textbf{52.7} \\
        \bottomrule
    \end{tabu}
    \caption{Results (\%) on \textbf{TimeQuestions}. The best results are in \textbf{bold} and the second bests are \underline{underlined}.}
    \label{tab:TimeQ}
\end{table}

\subsection{Main Results}
Table~\ref{tab:TempQ} and \ref{tab:TimeQ} report the results of compared methods on TempQuestions and TimeQuestions respectively.\footnote{The results of compared methods on TempQuestions are obtained from~\citeposs{JiaARSW18} \href{https://qa.mpi-inf.mpg.de/tequila/tequila.zip}{homepage} and the results of compared methods on TimeQuestions are provided by the authors of EXAQT~\citep{JiaPRW21}.} Our method, SF-TQA, achieves the best results on both two benchmarks. Specially, we improve the $F_1$ scores by $+3.6$ and $+7.1$ points on TempQuestions and TimeQuestions respectively. 
On TempQuestions, SF-TQA improves the Hit@1 and precision by $+4.3$ and $+1.9$ respectively. 
On TimeQuestions, SF-TQA achieves better recall ($5.2$ points higher) while EXAQT achieves better precision ($4.2$ points higher).
The reason could be the different strategies when dealing with unsolvable problems. EXAQT tends to output empty answers (which correspond to $1$ in precision) and SF-TQA degrades to the unrestricted generation of basic query graphs (which capture incomplete question meanings).  
The Hit@1 of SF-TQA is $2.6$ points lower than EXAQT might because EXAQT ranks all candidate answers while SF-TQA just randomly returns one candidate that satisfies the generated query.

\subsection{Ablation Studies}
We conduct ablations on the necessity of interpretations for intrinsic connections (i.e., IS-4 to 6). We analyze the result on questions with only relation constraints. The ablation results are illustrated in Table~\ref{tab:rc}. The 2nd row shows that without IS 4 to 6 the F$_1$ scores drop $1.5$ and $3.8$ points respectively on the benchmarks. The 3rd row shows that results obtained by generating only basic query graphs without any restriction will decrease the F$_1$ scores by $14.1$ and $4.7$ points respectively. The differences between the results on the two benchmarks might reflect the differences between underlying KGs. SF-TQA without IS 4 to 6 achieves acceptable results on TempQuestions, which might indicate that Freebase can provide sufficient temporal facts for comparisons. SF-TQA with only basic query graphs on TimeQuestions performs much better than on TempQuestions, which might indicate that Wikidata provides richer and finer relations between entities, thus the connections between mentioned entities and answers are more likely to be satisfied via simple relation paths.

\begin{table}[ht]
    \centering
    \begin{tabu} to \columnwidth  {X[4,l] X[1,r] X[1,r] X[1,r] X[1,r]}
        \toprule
        \rowfont[c]\bfseries \multirow{2}{*}{Method} & \multicolumn{2}{c}{TempQ.} & \multicolumn{2}{c}{TimeQ.}\\
        \cmidrule(lr){2-3}\cmidrule(lr){4-5}
        & H@1 & F$_1$ & H@1 & F$_1$ \\
        \midrule
        Full System & \textbf{37.5} & \textbf{38.1} & \textbf{41.3} & \textbf{40.7} \\
        \quad w/o IS-4 to 6 & \textbf{37.5} & \underline{36.6} & \underline{35.4} & \underline{36.9} \\
        \quad w/o IS & 29.2 & 24.0 & 34.7 & 36.0 \\
        \bottomrule
    \end{tabu}
    \caption{Results (\%) for the effectiveness of interpretation structures on relation constraints.}
    \label{tab:rc}
\end{table}

\subsection{Error Analysis}
We analyze the main errors of 100 questions of which the F$_1$ scores are less than $1$. The results are illustrated in Table~\ref{tab:errors}.

\begin{table}[h]
\begin{tabu} to \columnwidth {X[2.7,l] X[1,r] X[1,r]}
\toprule
\rowfont[c]\bfseries Main Error & TempQ. & TimeQ. \\
\midrule
Recognition Errors & 12\% & 36\% \\
Uncovered Constraints & 14\% & 20\% \\
Ranking Errors & 26\% & 2\% \\
Inconsistent Answers & 10\% & 8\% \\
Incomplete Knowledge & 38\% & 34\% \\
\bottomrule
\end{tabu}
\caption{The statistics of main errors of sampled questions.}
\label{tab:errors}
\end{table}
The 1st row counts the questions with incorrectly recognized entities or temporal constraints, which reveals that SF-TQA severely suffers from error propagations on TimeQuestions. The 2nd row counts the questions whose meaning can not be perfectly expressed by generated constraints (e.g., questions with multi-hop non-temporal property paths like \textit{``wife of the actor who played in the movie pinball wizard''}).
The 3rd row shows that our ranking model is hard to train with limited data (TempQuestions contains less than 1,000 training samples). Besides, data quality appears to be an important issue. In about 10\% of the sample questions, the provided answers are inconsistent with the knowledge in the given KG. For example, TimeQuestions annotate \textsf{2010\_F1\_Championship} (\href{https://www.wikidata.org/wiki/Q69934}{\url{wd:Q69934}}) as the answer of \textit{``Who won the 2010 f1 championship?''}. For over 1/3 of the sampled questions, KG can not provide sufficient evidence (e.g., occurrence times of the corresponding facts are not provided) for obtaining all answers.

\section{Related work}\label{sec:relwork}
\textbf{Temporal Information in Natural Language.} Temporal information has attracted the attention of AI and linguistics communities for a long time. 
% Supporting temporal intent in AI applications has attracted the attention of researchers since the last century~\citep{Bruce72}. 
Allen presents an interval-based temporal logic for reasoning the relation between time duration~\citep{Allen83} and a computation theory for action and time~\citep{Allen84}. He concludes 13 possible interval relations with their transitivity table.
\citet{ManiW00} introduces an annotation scheme for temporal expression in news and discusses its possible application in event ordering and event time alignment. 
TimeML~\citep{PustejovskyCISGSKR03,PustejovskyLBR10} is a specification for annotating temporal information from narratives. TimeML has become the de facto standard in the NLP community. It annotates time expressions, events, the relations between them, and the signals that trigger the relations in XML form. 

\textbf{Temporal KGQA.}
Early KGQA systems usually do not handle temporal constraints (e.g., \citealp{BerantL14}) or apply simple heuristics about their surface forms~\citep{BerantCFL13,BastH15,BaoDYZZ16}.
Some benchmarks that specifically focus on temporal intents in KGQA emerge in recent years, including TempQuestions~\citep{JiaARSW18}, TimeQuestions~\citep{JiaPRW21} and TempQA-WD~\citep{abs-2201-05793}. 
In terms of the technologies for temporal KGQA, \citet{JiaARSW18-2} proposes TEQUILA. It relies on limited hand-crafted rules to decompose complex temporal relations and solves composed simple questions via underlying general KGQA systems. 
EXAQT~\citep{JiaPRW21} uses Group Steiner Trees to anchor a KG sub-graph for each question, retrieving answers in the sub-graph with augmented temporal facts by an RGCN model.
Besides, there are also some researches specifically focus on question event-centric or temporal knowledge graphs. 
\citet{CostaGD20} proposes a question answering benchmark Event-QA over EventKG~\citep{GottschalkD18,GottschalkD19}.
\citet{SaxenaCT20} proposes CronQuestion over a sub-graph of Wikidata with a limited subset of relations for evaluating temporal KG embeddings.

In summary, existing temporal KGQA methods either analyze only the surface form of temporal constraints or rely on end-to-end neural models. While neural models might be robust to diversified representations of knowledge, they are hard to characterize the clear boundaries of temporal constraints (e.g., accurately filtering all events that occur before 2022). 

\textbf{KGQA via Query Graph Generation.} Constructing queries via exploring the relevant facts of mentioned entities is a common practice in KGQA, especially in the situations where only question-answers pairs are provided.
\citet{YihCHG15} defines query graphs that can be straightforwardly mapped to an executable logical query. They model the generation of query graphs as a staged search problem, where the query graphs are expanded by exploring legitimate predicate sequences starting from mentioned entities.
\citet{BaoDYZZ16} expands basic query graphs with 6 kinds of manually designed constraints including quantitative temporal and ordinal constraints.
\citet{LuoLLZ18} encodes query graphs of complex structures into a uniform vector representation for complex questions.
\citet{LanJ20} prunes the search space via early incorporation of constraints.

The existing query graph generation methods are not specifically designed for temporal constraints, they simply suppose that temporal or ordinal signals correspond to quantitative constraints.
Specifically, \citet{BaoDYZZ16}, \citet{LuoLLZ18}, and \citet{LanJ20} recognize time constraints via syntax signals and simply interpret them as general aggregation functions (e.g., greater than X, max at N), i.e., their interpretations of temporal constraints are similar to ''SF-TQA w/o IS-4 to 6'' (referring to Table \ref{tab:rc}).
In contrast, we systematically analyze the interpretation structure of temporal constraints, including the analysis of what kind of intrinsic connection can make events temporally related. 

\section{Conclusion and Future work}
In this paper, we study the logical constraint that corresponds to temporal intents in questions. Our main contributions can be summarized as follows:
\begin{itemize}
    \item We propose the idea of analyzing temporal intents via possible interpretation structures. We conclude the interpretation structures as SF-TCons, which allows one constraint expression has various interpretations.
    \item We propose the semantic-framework-based temporal question answering method, SF-TQA. SF-TQA mitigates the heterogeneity between expressions of temporal intents and KG facts. It enhances the query generation via structural restrictions provided by SF-TCons. 
    \item Our implementation of SF-TQA establishes new SOTAs on two benchmarks and improves the F$_1$ scores by $+3.6$ and $+7.1$ points respectively.
\end{itemize}

In the near future, it is worth exploring to alleviate the possible knowledge incompleteness in practical KG by developing a hybrid question-answering method on both knowledge graphs and web texts.
In addition, this paper focuses only on temporal intent, while problems in real configurations may contain both complex non-temporal and temporal intents. Therefore, it would be helpful to combine SF-TCons with general KGQA systems for complex questions.

\section*{Limitations}
\begin{itemize}
\item Due to the compositionality of natural language, a temporal question could be very complex, which is beyond the ability of our implemented QA system. For example, The following question is syntactically legitimate but can not be handled by SF-TQA: \textit{``What year did the second president of the United States, elected after the last spouse of the author of `Wish Tree for Washington, DC' was shot, marry his wife?''}
\item While the linguistic and entity annotations help SF-TQA alleviate the lack of structured supervision, they make it hard to apply SF-TQA to low-resource languages or questions with no named entities (e.g., \textit{``What are the important events that will happen at the turn of the century?''}). Besides, as a pipeline method, SF-TQA suffers from possible error propagations. 
\end{itemize}

\section*{Acknowledgements}
This work was supported by the National Natural Science Foundation of China (NSFC) under Grant No. 62072224. and the Program B for Outstanding PhD candidate of Nanjing University. The authors would like to thank all the participants of this work and anonymous reviewers.

% Entries for the entire Anthology, followed by custom entries

\bibliography{anthology,custom}
\bibliographystyle{acl_natbib}

\appendix
\section{Appendix for Different Training Strategies}\label{sec:app}
We also evaluated the effects of different training strategies as illustrated in Table~\ref{tab:train}. The results in 2 to 4 rows are obtained by simply using the irrelevant negatives, confusing negatives or randomly sampling negatives without classification respectively. The results show that both of the two types of negative sample are needed for training. The balanced sampling of the two types effectively improves SF-TQA on the smaller dataset, TempQuestions.

\begin{table}[ht]
    \centering
    \begin{tabu} to \columnwidth  {X[4,l] X[1,r] X[1,r] X[1,r] X[1,r]}
        \toprule
        \rowfont[c]\bfseries \multirow{2}{*}{Method} & \multicolumn{2}{c}{TempQ.} & \multicolumn{2}{c}{TimeQ.}\\
        \cmidrule(lr){2-3}\cmidrule(lr){4-5}
        & H@1 & F$_1$ & H@1 & F$_1$ \\
        \midrule
        Full System & \textbf{41.2} & \textbf{41.1} & \textbf{53.9} & \textbf{52.7} \\
        \quad w/o confusing & 34.9 & 35.9 & 49.5 & 49.3 \\
        \quad w/o irrelevant & 10.6 & 10.4 & 37.5 & 36.0 \\
        \quad random neg. & \underline{37.3} & \underline{37.4} & \underline{53.5} & \underline{52.6} \\
        \bottomrule
    \end{tabu}
    \caption{Results (\%) of different sampling strategies for the negative samples.}
    \label{tab:train}
\end{table}

\end{document}